\documentclass[11pt,a4paper]{article}
\RequirePackage{amsmath}
\usepackage{amsfonts}
\usepackage{amssymb}
\usepackage{xcolor}
\usepackage[hyperref]{acl2018}
\usepackage{times}
\usepackage{latexsym}
\usepackage{url}
\usepackage{float}
\usepackage{amsthm}
\usepackage{graphicx}
\usepackage{tikz}
\usepackage{enumitem}
\usetikzlibrary{arrows}
\usetikzlibrary{decorations.markings}

\newcommand{\todo}[1]{}

\newcommand{\tuple}[1]{\langle #1 \rangle}

\newcommand{\reals}{\mathbb{R}}

\newcommand{\up}[1]{^{(#1)}}
\newcommand{\down}[1]{_{(#1)}}

\newcommand{\hidden}[1]{}

\newcommand{\LL}{\mathcal{L}}

\newcommand{\grad}[2]{\frac{\partial #1}{\partial #2}}

\newcommand{\condProb}[2]{P\left(#1 \vert #2\right)}

\newcommand{\parGroup}[1]{\ensuremath{\rho^{(#1)}}}

\newcommand{\parAdapt}[1]{\ensuremath{\rho^{(#1)}}}

\newcommand{\parPrior}{\ensuremath{\pi}}

\usepackage{microtype}

\aclfinalcopy 
\def\aclpaperid{***} 

\title{Discovering User Groups for Natural Language Generation}

\author{Nikos Engonopoulos \and Christoph Teichmann \and Alexander Koller\\
Saarland University\\
\url{{nikos|cteichmann|koller}@coli.uni-saarland.de}
}
\date{}

\begin{document}

\maketitle

\begin{abstract}

  We present a model which predicts how individual users of a dialog system
  understand and produce utterances based on user groups. 
  In contrast to previous work, these user groups are not 
  specified beforehand, but learned in training. 
  We evaluate on two referring expression (RE) generation tasks;
  our experiments show that our model can identify user groups and learn
  how to most effectively talk to them, and can dynamically assign unseen users to
  the correct groups as they interact with the system.

\end{abstract}



\section{Introduction} \label{sec:introduction}

People vary widely both in their linguistic preferences when producing
language and in their ability to understand specific
natural-language expressions, depending on what they know about the
domain, their age and cognitive capacity, and many other factors. It
has long been recognized that effective NLG systems should 
therefore \emph{adapt} to the current user, in
order to generate language which works well for them. 
This adaptation needs to address all levels of the NLG pipeline,
including discourse planning \cite{paris88:_tailor}, sentence planning
\cite{walker07:_indiv_domain_adapt_senten_plann_dialog}, and RE generation
\cite{janarthanam14:_adapt_gener_dialog_system_using}, and depends on
many features of the user, including level of expertise and language
proficiency, age, and gender.

Existing techniques for adapting the output of an NLG system have shortcomings 
which limit their practical usefulness. Some systems need user-specific information 
in training \cite{ferreira14:_refer} and therefore cannot generalize to unseen 
users. Other systems assume that each user in the training data 
is annotated with their group, which allows them to learn a model 
from the data of each group.
However, hand-designed user groups may not reflect the true variability 
of the data, and may therefore inhibit the system's ability to flexibly adapt to new users.

In this paper, we present a user adaptation model for NLG systems
which induces user groups from training data in which these groups
were not annotated. At training time, we probabilistically assign users to groups and learn the language preferences for each group. At evaluation time, we assume that 
our system has a chance to interact with each new user repeatedly 
-- e.g., in the context of a dialogue system. It will then calculate 
an increasingly accurate estimate of the user's group membership
 based on observable behavior, and use it to generate 
utterances that are suitable to the user's true group.

We evaluate our model on two tasks involving the generation of referring 
expressions (RE). First, we predict the use of spatial relations in
 humanlike REs in the GRE3D domain~\cite{viethen2010speaker}
 using a log-linear production model in the spirit of \newcite{ferreira14:_refer}. 
Second, we predict the comprehension of generated REs, in a synthetic dataset
based on data from the GIVE Challenge domain \cite{StrDenGarGarKolThe11} 
with the log-linear comprehension model of \newcite{EngonopoulosVTK13}.
In both cases, we show that our model discovers user groups in the training 
data and infers the group of unseen users with high confidence after only a few 
interactions during testing. In the GRE3D domain, our system outperformed a strong baseline 
which used demographic information for the users.



\section{Related Work} \label{sec:related}

Differences between individual users have a
substantial impact on language comprehension. Factors that play a role
include level of expertise and spatial ability \cite{benyon1993developing}; 
age \cite{häuser17:_age}; gender \cite{navi-12}; or
language proficiency \cite{KolStrGarByrCasDalMooObe10}.

Individual differences are also reflected in the way people 
produce language. \newcite{viethen2008use} present a corpus study 
of human-produced REs (GRE3D3) for simple visual scenes, where they 
note two clearly distinguishable groups of speakers, one that always uses
a spatial relation and one that never does.
\newcite{ferreira14:_refer} show that 
a model using speaker-specific information
outperforms a generic model in predicting the attributes used by a speaker
when producing an RE. However, 
their system needs to have seen the particular speaker in training,
while our system can dynamically adapt to unseen users.
\newcite{ferreira2017improving} also demonstrate
that splitting speakers in predefined groups and training each group separately
improves the human likeness of REs compared to training individual user models.

The ability to adapt to the comprehension and production preferences of a user
is especially important in the context of a dialog system, where 
there are multiple chances of interacting with the same user. 
Some methods adapt to dialog system users 
by explicitly modeling the users' knowledge state.
An early example is \newcite{paris88:_tailor}; she selects a discourse plan for
a user, depending on their level of domain knowledge ranging between novice
and expert, but provides no mechanism for inferring the group to
which the user belongs. \newcite{rosenblum93:_partic_instr_dialog} try to
infer what knowledge a user possesses during dialogue, based on the
questions they ask. \newcite{janarthanam14:_adapt_gener_dialog_system_using}
 adapt to unseen users by using reinforcement learning with simulated users
to make a system able to adjust to the level of the user's knowledge.  
They use five predefined groups from which they generate the simulated users' behavior, 
but do not assign real users to these groups. 
Our system makes no assumptions about the user's knowledge and does not need to 
train with simulated users, or use any kind of information-seeking moves;
we instead rely on the groups that
are discovered in training and dynamically assign new, unseen users,
based only on their observable behavior in the dialog.

Another example of a user-adapting dialog component is SPaRKy
 \cite{walker07:_indiv_domain_adapt_senten_plann_dialog},
a trainable sentence planner that can tailor sentence plans to
individual users' preferences. This requires training on separate data
 for each user; in contrast to this, we leverage the similarities between users 
and can take advantage of the full training data.



\section{Log-linear models for NLG in dialog} \label{sec:basicModel}

We start with a basic model of the way in which people produce and comprehend language. In order to generalize over production and comprehension, we will simply say that a human language user exhibits a certain \emph{behavior} $b$ among a range of possible behaviors, in response to a \emph{stimulus} $s$. The behavior of a speaker is the utterance $b$ they produce in order to achieve a communicative goal $s$; the behavior of a listener is the meaning $b$ which they assign to the utterance $s$ they hear.

Given this terminology, we define a basic log-linear model~\cite{BergerPP96} of language use as follows:
\begin{eqnarray} \label{eq:basicModel}
	\condProb{b}{s;\rho} = \displaystyle \frac{\exp(\rho \cdot \phi(b,s))}{\sum_{b'}{\exp(\rho \cdot  \phi(b', s))}}
\end{eqnarray}

\noindent
where $\rho$ is a real-valued parameter vector of length $n$
and $\phi(b,s)$ is a vector of real-valued \textit{feature functions} $f_1,...,f_n$ over behaviors and stimuli. The parameters can be trained by maximum-likelihood estimation from a corpus of observations $(b,s)$. In addition to maximum-likelihood training it is possible to include some prior probability distribution, which expresses our belief about the probability of any parameter vector and which is generally used for regularization. The latter case is referred to as \emph{a posteriori} training, which selects the value of $\rho$ that maximizes the product of the parameter probability and the probability of the data.

\begin{figure}
\centering
\includegraphics[width=0.75\columnwidth]{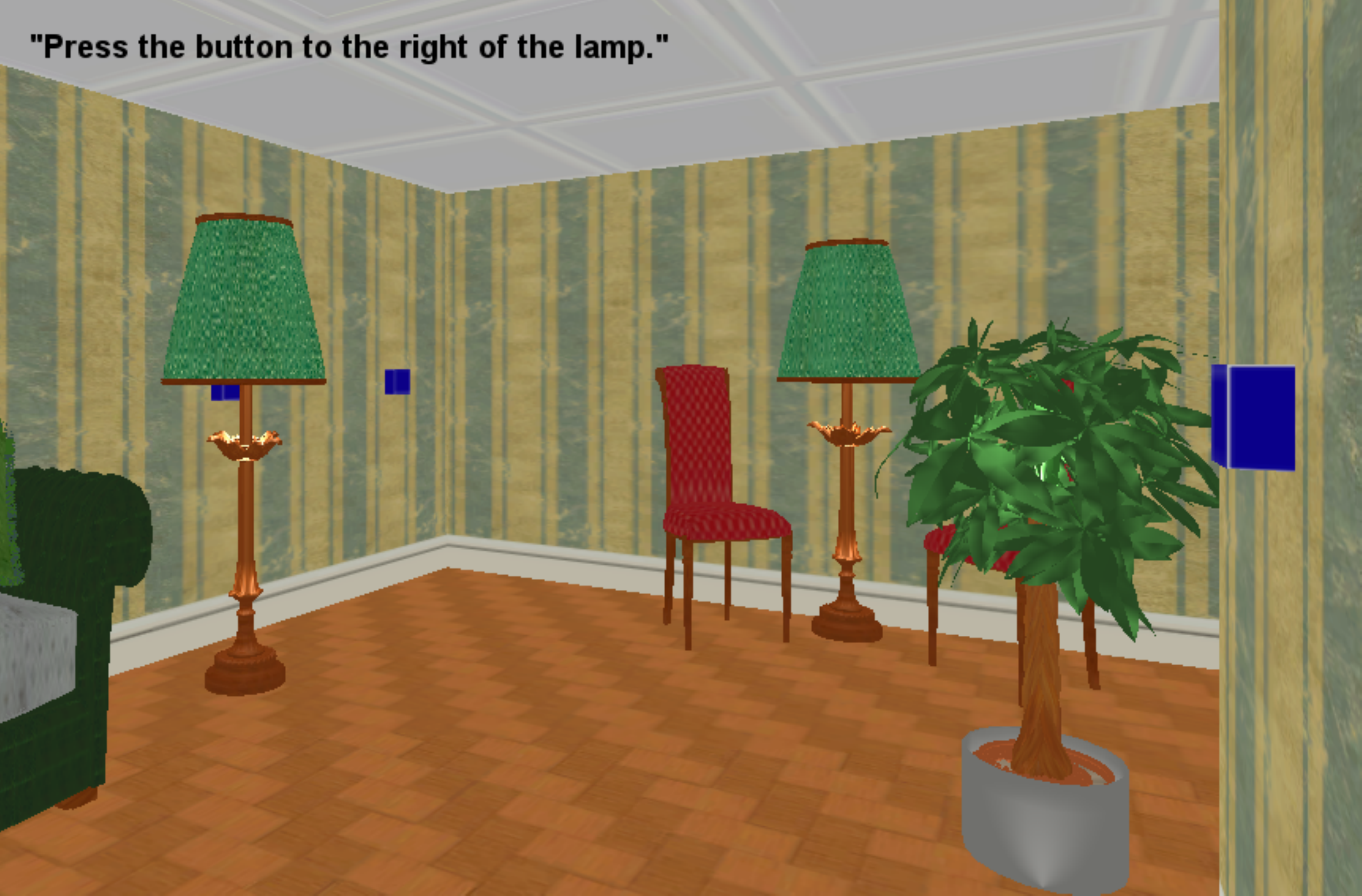}
\caption{A visual scene and a system-generated instruction from the GIVE challenge.} 
\label{fig:GIVE}
\end{figure}

In this paper, we focus on the use of such models in the context of the NLG module of a dialogue system, and more specifically on the generation of referring expressions (REs). Using \eqref{eq:basicModel} as a \emph{comprehension} model, \newcite{EngonopoulosVTK13} developed an RE generation model in which the stimulus $s=(r,c)$ consists of an RE $r$ and a visual context $c$ of the GIVE Challenge \cite{StrDenGarGarKolThe11}, as illustrated in Fig.~\ref{fig:GIVE}. The behavior is the object $b$ in the visual scene to which the user will resolve the RE. Thus for instance, when we consider the RE $r=$``the blue button'' in the context of Fig.~\ref{fig:GIVE}, the log-linear model may assign a higher probability to the button on the right than to the one in the background. \newcite{EngonopoulosK14} develop an algorithm for generating the RE $r$ which maximizes $\condProb{b^*}{s;\rho}$, where $b^*$ is the intended
referent in this setting.

Conversely, log-linear models can also be used to directly capture how a human speaker would refer to an object in a given scene. In this case, the stimulus $s = (a,c)$ consists of the target object $a$ and the visual context $c$, and the behavior $b$ is the RE. We follow \newcite{ferreira14:_refer} in training individual models for the different attributes which can be used in the RE (e.g., that $a$ is a button; that it is blue; that the RE contains a binary relation such as ``to the right of''), such that we can simply represent $b$ as a binary choice $b \in \{1,-1\}$ between whether a particular attribute should be used in the RE or not. We can then implement an analog of Ferreira's model in terms of \eqref{eq:basicModel} by using feature functions $\phi(b,a,c) = b \cdot \phi'(a,c)$, where $\phi'(a,c)$ corresponds to their \textit{context} features, which do not capture any speaker-specific information.



\section{Log-linear models with user groups} \label{sec:augmentation}

As discussed above, a user-agnostic model such as \eqref{eq:basicModel} does not do justice to the variability of language comprehension and production across different speakers and listeners. We will therefore extend it to a model which distinguishes different \emph{user groups}. We will not try to model why\footnote{E.g., in the sense of explicitly modeling sociolects or the difference between novice system users vs.\ experts.} users behave differently. Instead our model sorts users into groups simply based on the way in which they respond to stimuli, in the sense of Section~\ref{sec:basicModel}, and implements this by giving each group $g$ its own parameter vector $\parAdapt{g}$. As a theoretical example, Group 1 might contain users who reliably comprehend REs which use colors (``the green button''), whereas Group 2 might contain users who more easily understand relational REs (``the button next to the lamp''). These groups are then discovered at training time.

When our trained NLG system starts interacting with an unseen user $u$, it will infer the group to which $u$ belongs based on $u$'s observed responses to previous stimuli. Thus as the dialogue with $u$ unfolds, the system will have an increasingly precise estimate of the group to which $u$ belongs, and will thus be able to generate language which is increasingly well-tailored to this particular user.

\subsection{Generative story}

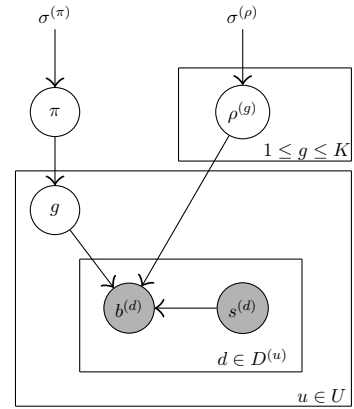
\begin{figure} 
\begin{center}
	\begin{tikzpicture}[scale=0.65, every node/.style={transform shape}]

\node (varPrior) at (0,0) {$\sigma\up{\parPrior}$};

\node (varAdapt) at (3.8,0) {$\sigma\up{\rho}$};

\node[circle,draw, minimum size=1cm] (prior) at (0,-2) {$\parPrior$};

\node[circle,draw, minimum size=1cm] (adapt) at (3.8,-2) {$\parAdapt{g}$};

\node[circle,draw, minimum size=1cm] (member) at (0,-4) {$g$};

\node[circle,draw, minimum size=1cm,fill=gray!60] (behavior) at (1.5,-6) {$b\up{d}$};

\node[circle,draw, minimum size=1cm,fill=gray!60] (stimulus) at (3.8,-6) {$s\up{d}$};

\node (aBoxLT) at (2.5,-1.1) {};
\node (aBoxRT) at (6,-1.1) {};
\node (aBoxLB) at (2.5,-3) {};
\node (aBoxRB) at (6,-3) {};
\node (aBoxCount) at (5.1,-2.8) {$1 \leq g \leq K$};

\draw (aBoxLT.center) -- (aBoxRT.center);
\draw (aBoxLB.center) -- (aBoxRB.center);
\draw (aBoxLT.center) -- (aBoxLB.center);
\draw (aBoxRB.center) -- (aBoxRT.center);

\node (uBoxLT) at (-0.8,-3.2) {};
\node (uBoxRT) at (6,-3.2) {};
\node (uBoxLB) at (-0.8,-8) {};
\node (uBoxRB) at (6,-8) {};
\node (uBoxCount) at (5.4,-7.8) {$u \in U$};

\draw (uBoxLT.center) -- (uBoxRT.center);
\draw (uBoxLB.center) -- (uBoxRB.center);
\draw (uBoxLT.center) -- (uBoxLB.center);
\draw (uBoxRB.center) -- (uBoxRT.center);

\node (oBoxLT) at (0.5,-5) {};
\node (oBoxRT) at (5,-5) {};
\node (oBoxLB) at (0.5,-7.3) {};
\node (oBoxRB) at (5,-7.3) {};
\node (oBoxCount) at (4,-7) {$d \in D\up{u}$};

\draw (oBoxLT.center) -- (oBoxRT.center);
\draw (oBoxLB.center) -- (oBoxRB.center);
\draw (oBoxLT.center) -- (oBoxLB.center);
\draw (oBoxRB.center) -- (oBoxRT.center);

\draw[decoration={markings,mark=at position 1 with {\arrow[scale=2]{>}}}, postaction={decorate}] (varPrior) -- (prior);

\draw[decoration={markings,mark=at position 1 with {\arrow[scale=2]{>}}}, postaction={decorate}] (prior) -- (member);

\draw[decoration={markings,mark=at position 1 with {\arrow[scale=2]{>}}}, postaction={decorate}] (varAdapt) -- (adapt);

\draw[decoration={markings,mark=at position 1 with {\arrow[scale=2]{>}}}, postaction={decorate}] (adapt) -- (behavior);

\draw[decoration={markings,mark=at position 1 with {\arrow[scale=2]{>}}}, postaction={decorate}] (member) -- (behavior);

\draw[decoration={markings,mark=at position 1 with {\arrow[scale=2]{>}}}, postaction={decorate}] (stimulus) -- (behavior);

\end{tikzpicture}
\end{center}
\caption{Plate diagram for the user group model.} \label{fig:plates_drawn}
\end{figure}

We assume training data $D = \{(b_i,s_i,u_i)\}_i$ which contains stimuli $s_i$ together with the behaviors $b_i$ which the users $u_i$ exhibited in response to $s_i$. We write $D\up{u} = \{(b^u_1, s^u_1), \ldots (b^u_N, s^u_N)\}$ for the data points for each user $u$.

The generative story we use is illustrated in Fig.~\ref{fig:plates_drawn}; observable variables are shaded gray, unobserved variables and parameters to be set in training are shaded white and externally set hyperparameters have no circle around them. Arrows indicate which variables and parameters influence the probability distribution of other variables.

We assume that each user belongs to a group $g \in \{1,\ldots,K\}$, where the number $K$ of groups is fixed beforehand based on, e.g., held out data. A group $g$ is assigned to $u$ at random from the distribution
\begin{eqnarray} \label{eq:prior}
\condProb{g}{\parPrior} = \displaystyle
\frac{\exp(\parPrior_g)}{\sum_{g'=1}^K \exp(\parPrior_{g'})}
\end{eqnarray}
Here $\parPrior \in \reals^K$ is a vector of weights, which defines how probable each group is a-priori.

\begin{figure*}
	\begin{eqnarray}
		P(D;\theta) = & \displaystyle \left(\prod_{u \in U} \sum_{g =1}^K P(g|\parPrior) \cdot \prod_{d \in D\up{u}} \condProb{b_d}{s_d;\parGroup{g}}\right) \cdot \left(\mathcal{N}\left(\parPrior|0,\sigma\up{\parPrior}\right) \cdot \prod_{g=1}^K\mathcal{N}\left(\parAdapt{g}|0,\sigma\up{\rho}\right) \right) \label{eqn::probability} \\
		\LL(\theta) = & \displaystyle \sum_{u \in U} \log \sum_{g = 1}^K P(g|\parPrior) \cdot \prod_{d \in D\up{u}} \condProb{b_d}{s_d;\parGroup{g}} \label{eqn::loglike} \\
		\mathcal{AL}(\theta) = & \displaystyle \sum_{u \in U} \sum_{g = 1}^K \left( \condProb{g}{D\up{u};\theta\down{i-1}} \cdot \left( \log P(g|\parPrior) + \sum_{d \in D_u} \log \condProb{b_d}{s_d;\parGroup{g}}\right)\right) \label{eqn::grad}
	\end{eqnarray}
\end{figure*}

We replace the single parameter vector $\rho$ of \eqref{eq:basicModel} with group-specific parameters vectors $\parAdapt{g}$, thus obtaining a potentially different log-linear model $\condProb{b}{s; \parGroup{g}}$ for each group. After assigning a group, our model generates responses $b^u_1,\ldots,b^u_N$ at random from $\condProb{b}{s; \parAdapt{g}}$, based on the group specific parameter vector and the stimuli $s^u_1,\ldots,s^u_N$. This accounts for the generation of the data.

We model the parameter vectors $\parPrior \in \reals^K$, and $\parAdapt{g} \in \reals^n$ for every $1 \leq g \leq K$ as drawn from normal distributions $\mathcal{N}(0,\sigma\up{\pi})$, and $\mathcal{N}(0,\sigma\up{\rho})$, which are centered at $0$ with externally given variances and no covariance between parameters. This has the effect of making parameter choices close to zero more probable. Consequently, our models are unlikely to contain large weights for features that only occurred a few times or which are only helpful for a few examples. This should reduce the risk of overfitting the training set.

The equation for the full probability of the data and a specific parameter setting is given in \eqref{eqn::probability}. The left bracket contains the likelihood of the data, while the right bracket contains the prior probability of the parameters.

\subsection{Predicting user behavior}
\label{sec:group-posterior}

Once we have set values $\theta = (\parPrior, \parAdapt{1}, \ldots, \parAdapt{K})$ for all the parameters, we want to predict what behavior $b$ a user $u$ will exhibit in response to a stimulus $s$. If we encounter a completely new user $u$, the prior user group distribution from \eqref{eq:prior} gives the probability that this user belongs to each group. We combine this with the group-specific log-linear behavior models to obtain the distribution:
\begin{eqnarray} \label{eq:group-loglin}
\condProb{b}{s; \theta} = 
 \sum_{g = 1}^K 
 \condProb{b}{s; \parGroup{g}} \cdot \condProb{g}{\parPrior}
 \label{expr::findREUser} \end{eqnarray}

\noindent
Thus, we have a group-aware replacement for \eqref{eq:basicModel}.

Furthermore, in the interactive setting of a dialogue system, we may have multiple opportunities to interact with the same user $u$. We can then develop a more precise estimate of $u$'s group based on their responses to previous stimuli. Say that we have made the previous observations $D\up{u} = \{\tuple{s_1,b_1},\dots,\tuple{s_N,b_N}\}$ for user $u$. Then we can use Bayes' theorem to calculate a \emph{posterior} estimate for $u$'s group membership:
\begin{eqnarray} \label{eq:posterior}
\condProb{g}{D\up{u}; \theta}
\propto
\condProb{D\up{u}}{\parGroup{g}} \cdot \condProb{g}{\parPrior}
\end{eqnarray}

This posterior balances whether a group is likely in general against whether members of that group behave as $u$ does. We can use $P_u(g) = \condProb{g}{D\up{u}; \theta}$ as our new estimate for the group membership probabilities for $u$ and replace \eqref{eq:group-loglin} with:
\begin{eqnarray}  \label{eq:posterior-loglin}
\condProb{b}{s, D\up{u}; \theta} = 
 \sum_{g = 1}^K \condProb{b}{s; \parGroup{g}} \cdot P_u(g)
 \end{eqnarray}

\noindent for the next interaction with $u$.

An NLG system can therefore adapt to each new user over time. Before the first interaction with $u$, it has no specific information about $u$ and models $u$'s behavior based on \eqref{eq:group-loglin}. As the system interacts with $u$ repeatedly, it collects observations $D\up{u}$ about $u$'s behavior. This allows it to calculate an increasingly accurate posterior $P_u(g) = \condProb{g}{D\up{u}; \theta}$ of $u$'s group membership, and thus generate utterances which are more and more suitable to $u$ using \eqref{eq:posterior-loglin}.

\section{Training} \label{sec::training}

So far we have not discussed how to find settings for the parameters $\theta = \parPrior,\parAdapt{1},\dots,\parAdapt{K}$, which define our probability model. The key challenge for training is the fact that we want to be able to train while treating the assignment of users to groups as unobserved.

We will use a maximum \emph{a posteriori} estimate for $\theta$, i.e., the setting which maximizes \eqref{eqn::probability} when $D$ is our training set. We will first discuss how to pick parameters to maximize only the left part of \eqref{eqn::probability}, i.e., the data likelihood, since this is the part that involves unobserved variables. We will then discuss handling the parameter prior in section \ref{sec::prior}.

\subsection{Expectation Maximization}

Gradient descent based methods \cite{NocedalW06} exist for finding the parameter settings which maximize the likelihood for log-linear models, under the conditions that all relevant variables are observed in the training data. If group assignments were given, gradient computations, and therefore gradient based maximization, would be straightforward for our model. One algorithm specifically designed to solve maximization problems with unknown variables by reducing them to the case where all variables are observed, is the expectation maximization (EM) algorithm \cite{NealH99}. Instead of maximizing the data likelihood from \eqref{eqn::probability} directly, EM equivalently maximizes the log-likelihood, given in \eqref{eqn::loglike}. It helps us deal with unobserved variables by introducing ``pseudo-observations'' based on the expected frequency of the unobserved variables.

EM is an iterative algorithm which produces a sequence of parameter settings $\theta\down{1},\dots,\theta\down{n}$. Each will achieve a larger value for (\ref{eqn::loglike}). Each new setting is generated in two steps: (1) an lower bound on the log-likelhood is generate and (2) the new parameter setting is found by optimizing this lower bound. To find the lower bound we compute the probability for every possible value the unobserved variables could have had, based on the observed variables and the parameter setting $\theta\down{i-1}$ from the last iteration step.  Then the lower bound  essentially assumes that each assignment was seen with a frequency equal to these probabilities - these are the ``pseudo-observations''.

In our model the unobserved variables are the assignments of users to groups. The probability of seeing each user $u$ assigned to a group, given all the data $D\up{u}$ and the model parameters from the last iteration $\theta\down{i-1}$, is simply the posterior group membership probability $\condProb{g}{D\up{u};\theta\down{i-1}}$. The lower bound is then given by \eqref{eqn::grad}. This is the sum of the log probabilities of the data points under each group model, weighted by $\condProb{g}{D\up{u};\theta\down{i-1}}$. We can now use gradient descent techniques to optimize this lower bound.

\subsubsection{Maximizing the Lower Bound}

To fully implement EM we need a way to maximize \eqref{eqn::grad}. This can be achieved with gradient based methods such as L-BFGS \cite{NocedalW06}. Here the gradient refers to the vector of all partial derivatives of the function with respect to each dimension of $\theta$. We therefore need to calculate these partial derivatives.

There are existing implementations of the gradient computations our base model such as in \newcite{EngonopoulosVTK13}. The gradients of \eqref{eqn::grad} for each of the $\parAdapt{g}$ is simply the gradient for the base model on each datapoint $d$ weighted by $\condProb{g}{D\up{u};\theta\down{i-1}}$ if $d \in D_u$, i.e., the probability  that the user $u$ from which the datapoint originates belongs to group $g$. We can therefore compute the gradients needed for each $\parAdapt{g}$ by using implementations developed for the base model.

We also need gradients for the parameters in $\parPrior$, which are only used in our extended model. We can use the rules for computing derivatives to find, for each dimension $g$:

\begin{equation*}
\grad{\mathcal{UL}(\theta)}{\parPrior_g} = \displaystyle \sum_{u \in U} P_u(g) - \frac{\exp\left({\parPrior}_g\right)}{\sum_{g' = 1}^K \exp\left({\parPrior}_{g'}\right)}
\end{equation*}

where $P_u(g) = \condProb{g}{D\up{u};\theta\down{i-1}}$. Using these gradients we can use L-BFGS to maximize the lower bound and implement the EM iteration.

\subsection{Handling the Parameter Prior} \label{sec::prior}

So far we have discussed maximization only for the likelihood without accounting for the prior probabilities for every parameter. To obtain our full training objective we add the log of the right hand side of \eqref{eqn::probability}:

\begin{equation}
\log \left(\mathcal{N}\left(\parPrior|0,\sigma\up{\parPrior}\right) \cdot \prod_{g=1}^K\mathcal{N}\left(\parAdapt{g}|0,\sigma\up{\rho}\right) \right) \nonumber
\end{equation}

i.e., the parameter prior, to \eqref{eqn::loglike} and \eqref{eqn::grad}. The gradient contribution from these priors can be computed with standard techniques.

\subsection{Training Iteration}

We can now implement an EM loop, which maximizes \eqref{eqn::probability} as follows: we randomly pick an initial value $\theta\down{0}$ for all parameters. Then we repeatedly compute the $\condProb{g}{D\up{u};\theta\down{i-1}}$ values and maximize the lower bound using L-BFGS to find $\theta\down{i}$. This EM iteration is guaranteed to eventually converge towards a local optimum of our objective function. Once change in the objective falls below a pre-defined threshold, we keep the final $\theta$ setting.

For our implementation we make a small improvement to the approach: L-BFGS is itself an iterative algorithm and instead of running it until convergence every time we need to find a new $\theta\down{i}$, we only let it take a few steps. Even if we just took a single L-BFGS step in each iteration, we would still obtain a correct algorithm \cite{NealH99} and this has the advantage that we do not spend time trying to find a $\theta\down{i}$ which is a good fit for the likely poor group assignments $\condProb{g}{D\up{u};\theta\down{i-1}}$ we obtain from early parameter estimates.


\section{Evaluation} \label{sec:evaluation}

Our model can be used in any component of a dialog system for which a prediction
of the user's behavior is needed. In this work, we evaluate it in two
NLG-related prediction tasks: RE production and RE comprehension. In both cases
we evaluate the ability of our model to predict the user's behavior given a
stimulus. We expect our user-group model to gradually improve its prediction accuracy
compared to a generic baseline without user groups as it sees more observations
from a given user.

In all experiments described below we set the prior variances 
$\sigma_{\gamma}=1.0$ and $\sigma_{\pi}=0.3$ after trying out values between
0.1 and 10 on the training data of the comprehension experiment.

\subsection{RE production}\label{sub:production}

\paragraph{Task} The task of RE generation can be split in two
steps: \emph{attribute selection}, the selection of the visual attributes
to be used in the RE such as color, size, relation to other objects and
\emph{surface realization}, the generation of a full natural language expression. 
We focus here on attribute selection: given a visual scene and a target object, 
we want to predict the set of attributes of the target object that a human
speaker would use in order to describe it. Here we treat attribute selection in
terms of individual classification decisions on whether to use each attribute,
as described in Section~\ref{sec:basicModel}.
More specifically, we focus on predicting whether the speaker will use a \emph{spatial
relation} to another object (``landmark''). Our motivation for choosing this attribute 
stems from the fact that previous authors ~\cite{viethen2008use,ferreira14:_refer}
have found substantial variation between different users with respect to their preference
towards using spatial relations. 

\begin{figure} 
\centering
\includegraphics[width=0.6\columnwidth]{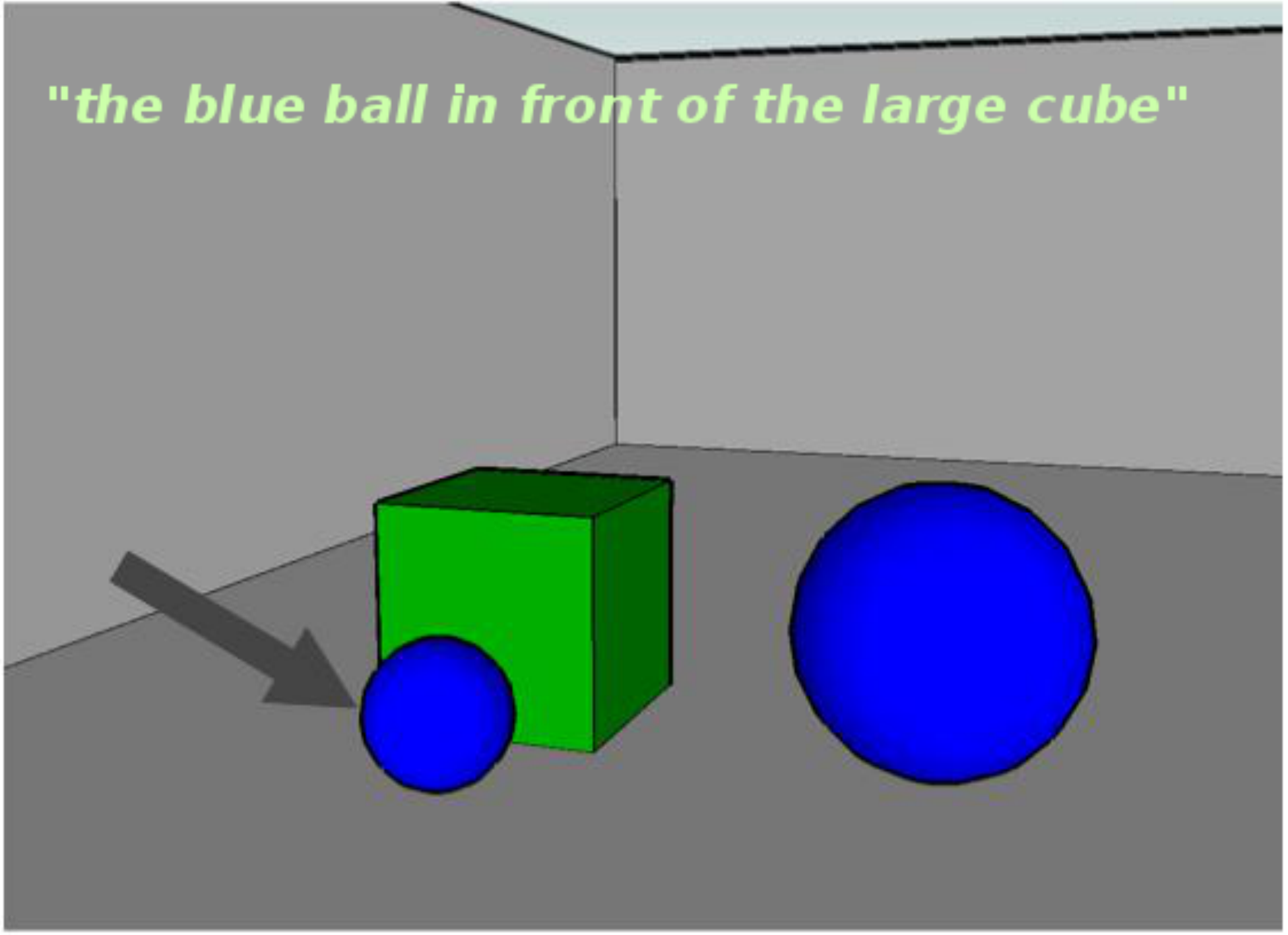} 
\caption{A sample scene with a human-produced RE from the GRE3D3 dataset.} 
\label{fig:gre3d} 
\end{figure}

\paragraph{Data} We use the GRE3D3 dataset of human-produced
REs~\cite{viethen2010speaker}, which contains 630 descriptions for 10 scenes
collected from 63 users, each describing the same target object in each scene.
$35\%$ of the descriptions in this corpus use a spatial relation. An example of
such a scene can be seen in Fig.~\ref{fig:gre3d}.

\paragraph{Models}

We use two baselines for comparison: 

\begin{itemize}[leftmargin=0pt]
\item[] \emph{Basic}: The state-of-the-art model on this task with this dataset, 
under the assumption that users are seen in training, is
presented in \newcite{ferreira14:_refer}. They define context features such as
type of relation between the target object and its landmark, number of object of
the same color or size, etc., then train an SVM classifier to predict the use of each
attribute. We recast their model in terms of a log-linear model with the same features,
to make it fit with the setup of Section~\ref{sec:basicModel}.

\item[] \emph{Basic++}: \newcite{ferreira14:_refer} also take speaker features into account.
We do not use speaker identity and the speaker's attribute frequency vector, 
because we only evaluate on unseen users. 
We do use their other speaker features (age, gender), 
together with \emph{Basic}'s context features; 
this gives us a strong baseline which is aware of manually
annotated user group characteristics.
\end{itemize}

We compare these baselines to our \emph{Group}  model 
for values of $K$ between 1 and 10, using the exact same features as 
\emph{Basic}. We do not use the speaker features of \emph{Basic++}, 
because we do not want to rely on manually annotated groups. 
Note that our results are not directly comparable with those 
of \newcite{ferreira14:_refer}, because of a different training-test split:
their model requires having seen speakers in training, 
while we explicitly want to test our model's ability to generalize to unseen users.

\paragraph{Experimental setup} We evaluate using cross-validation, splitting the
folds so that all speakers we see in testing are previously unseen
in training. We use 9 folds in order to have folds of the same size (each
containing 70 descriptions coming from 7 speakers). At each iteration we train
on 8 folds and test on the 9th. At test time, we process each test instance
iteratively: we first predict for each instance whether the user $u$ would use a
spatial relation or not and test our prediction; we then add the actual
observation from the corpus to the set $D\up{u}$ of observations for this particular
user, in order to update our estimate about their group membership.

\paragraph{Results}

\begin{figure}
\centering
 \includegraphics[width=1.0\columnwidth]{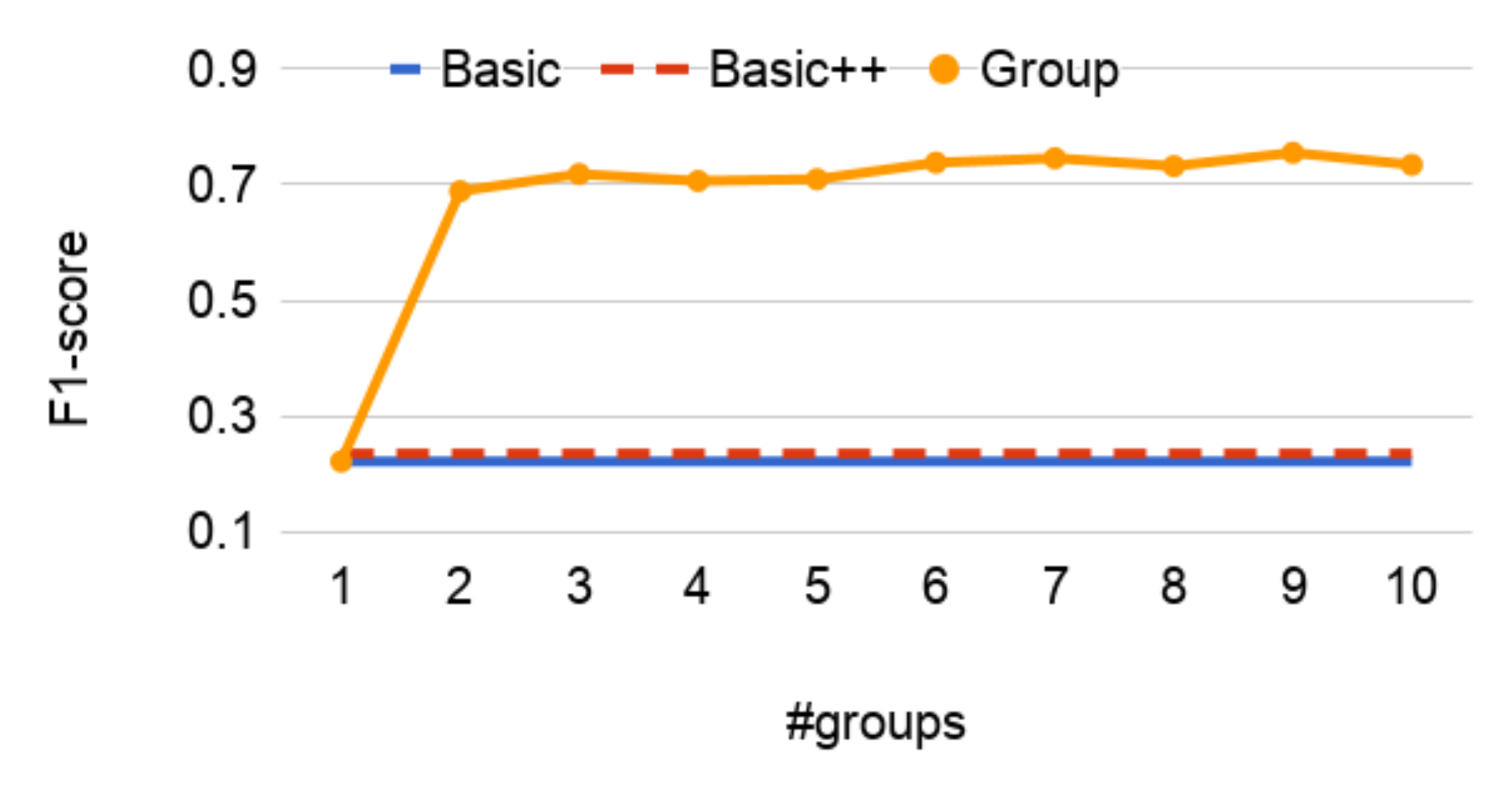}
\vspace{-1cm}
\caption{F1 scores on test data for values of $K$ between $1$ and $10$ in the production experiment.}
\label{fig:gre3d3-f1} \end{figure}

Figure~\ref{fig:gre3d3-f1} shows the test F1-score (micro-averaged over all
folds) as we increase the number of groups, compared to the baselines. 
For our \emph{Group} models, these are averaged over all interactions with the user.
Our model gets F1-scores between $0.69$ and $0.76$ for all values of $K>1$, 
outperforming both \emph{Basic} ($0.22$) and \emph{Basic++} ($0.23$).

\begin{figure}

\centering

\includegraphics[width=1.0\columnwidth]{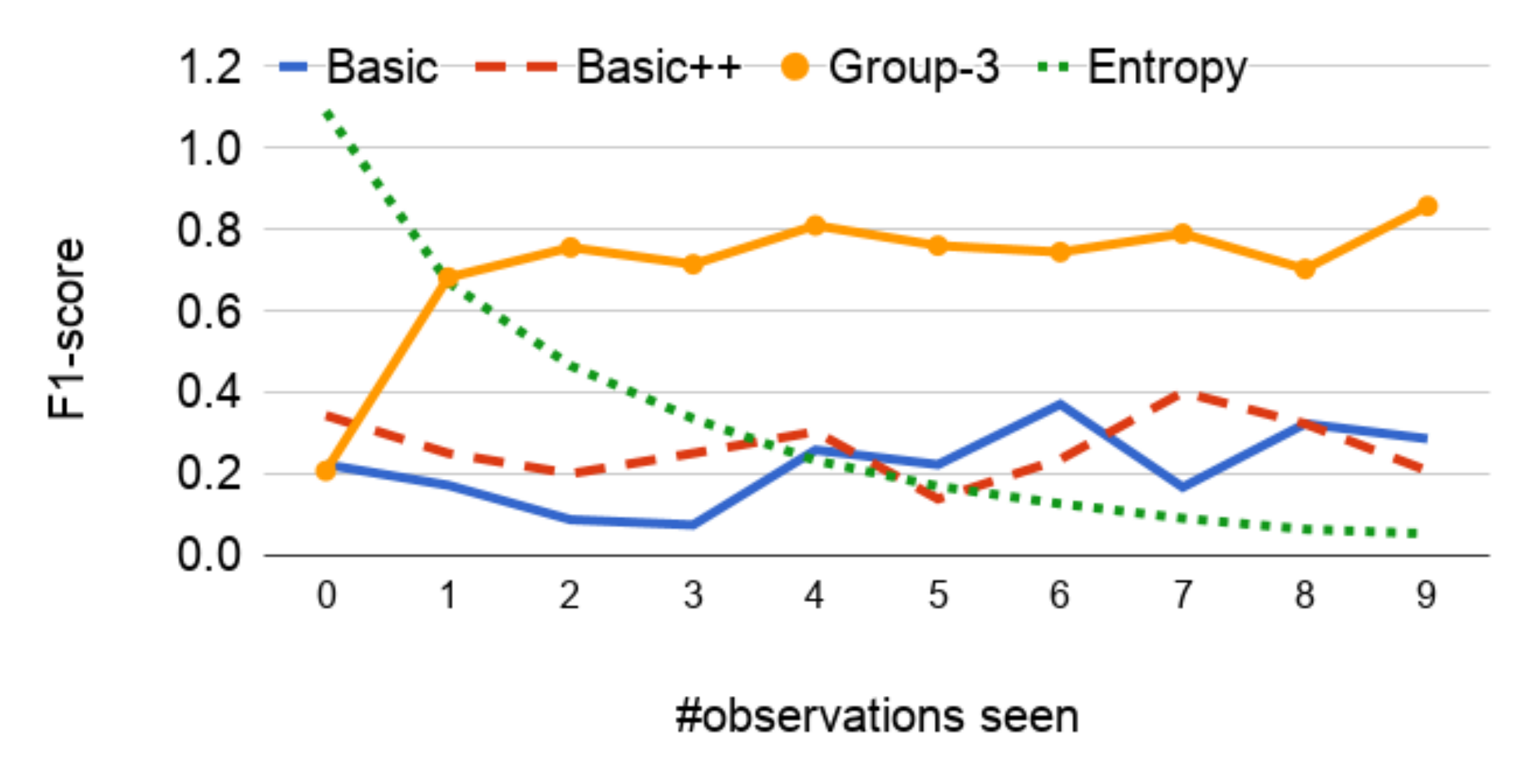} 
\vspace{-1cm}
\caption{F1-score evolution with increasing number of observations from the user in the production experiment.} 
\label{fig:gre3d3-time} 
\end{figure}

In order to take a closer look at our model's behavior, we also show the
accuracy of our model as it observes more instances at test time. We compare the
model with $K=3$ groups against the two baselines.
Figure~\ref{fig:gre3d3-time} shows that the group model's F1-score increases dramatically after
the first two observations and then stays high throughout the test phase, 
always outperforming both baselines by at least 0.37 F1-score points after the 
first observation. The baseline models of course are not expected to improve 
with time; fluctuations are due to differences between the visual scenes.
In the same figure, we plot the evolution of the entropy of the group model's posterior
distribution over the groups (see (\ref{eq:posterior})). As expected, the model is 
highly uncertain at the beginning of the test phase about which group the user 
belongs to, then gets more and more certain as the set $D\up{u}$ of observations  
from that user grows.

\subsection{RE comprehension}\label{sub:comprehension}
 
\paragraph{Task}
Our next task is to predict the referent to which a user will resolve an RE in
the context of a visual scene. Our model is given a stimulus $s=(r,
c)$ consisting of an instruction containing an RE $r$ and a visual context
$c$ and outputs a probability distribution over all possible referents $b$. Such
a model can be used by a probabilistic RE generator to select an RE which is
highly likely to be correctly understood by the user or predict potential
misunderstandings (see Section~\ref{sec:basicModel}). 

\paragraph{Data}
We use the GIVE-2.5 corpus for training and the GIVE-2 corpus for 
testing our model (the same used by \newcite{EngonopoulosVTK13}). These contain recorded
observations of dialog systems giving instructions to users who play a game in a
3D environment. Each instruction contains an RE $r$, which is recorded in the data
together with the visual context $c$ at the time the instruction was given. The object
$b$ which the user understood as the referent of the RE is inferred by the immediately subsequent
action of the user. In total, we extracted 2927 observations by 403 users from GIVE-25 and 
5074 observations by 563 users from GIVE-2. 

\paragraph{Experimental setup} 
We follow the training method described in
Section~\ref{sec:basicModel}. 
At test time, we present the observations from each user in the order they occur 
in the test data; for each stimulus, we ask our models to predict the referent $a$ which the user 
understood to be the referent of the RE, and compare with the recorded observation.
We subsequently add the recorded observation to the dataset for the user and continue.

\paragraph{Models}

As a baseline, we use the \emph{Basic} model described in Section
\ref{sec:basicModel}, with the features of the ``semantic'' model 
of \newcite{EngonopoulosVTK13}. Those features capture
information about the objects in the visual scene (e.g. salience) 
and some basic semantic properties of the RE (e.g. color, position). 
We use those features for our \emph{Group} model as well, and evaluate for 
$K$ between 1 and 10.

\paragraph{Results on GIVE data}

\emph{Basic} had a test accuracy of 72.70\%, which was almost identical
with the accuracy of our best \emph{Group} model for $K=6$ (72.78\%).
This indicates that our group model does not differentiate between users. 
Indeed, after training, the 6-group model assigns
$81\%$ prior probability to one of the groups, and effectively gets stuck with this
assignment while testing; the mean entropy of the posterior group distribution only falls from 
an initial 1.1 to 0.7 after 10 observations. 

We speculate that the reason behind this is that the
features we use are not sensitive enough to capture the differences between the users in
this data. Since our model relies completely on observable behavior, 
it also relies on the ability of the features to make relevant distinctions between users.

\paragraph{Results on synthetic data}

In order to test this hypothesis, we made a synthetic dataset based on the GIVE datasets 
with 1000 instances from 100 users, in the following way:
for each user, we randomly selected 10 scenes from GIVE-2, and replaced the target
the user selected, so that half of the users always select the target 
with the highest visual salience, and the other half always select the one with the lowest. 
Our aim was to test whether our model is capable of identifying groups when they are
clearly present in the data and exhibit differences which our
features are able to capture.

\begin{figure}
\centering
\includegraphics[width=1.0\columnwidth]{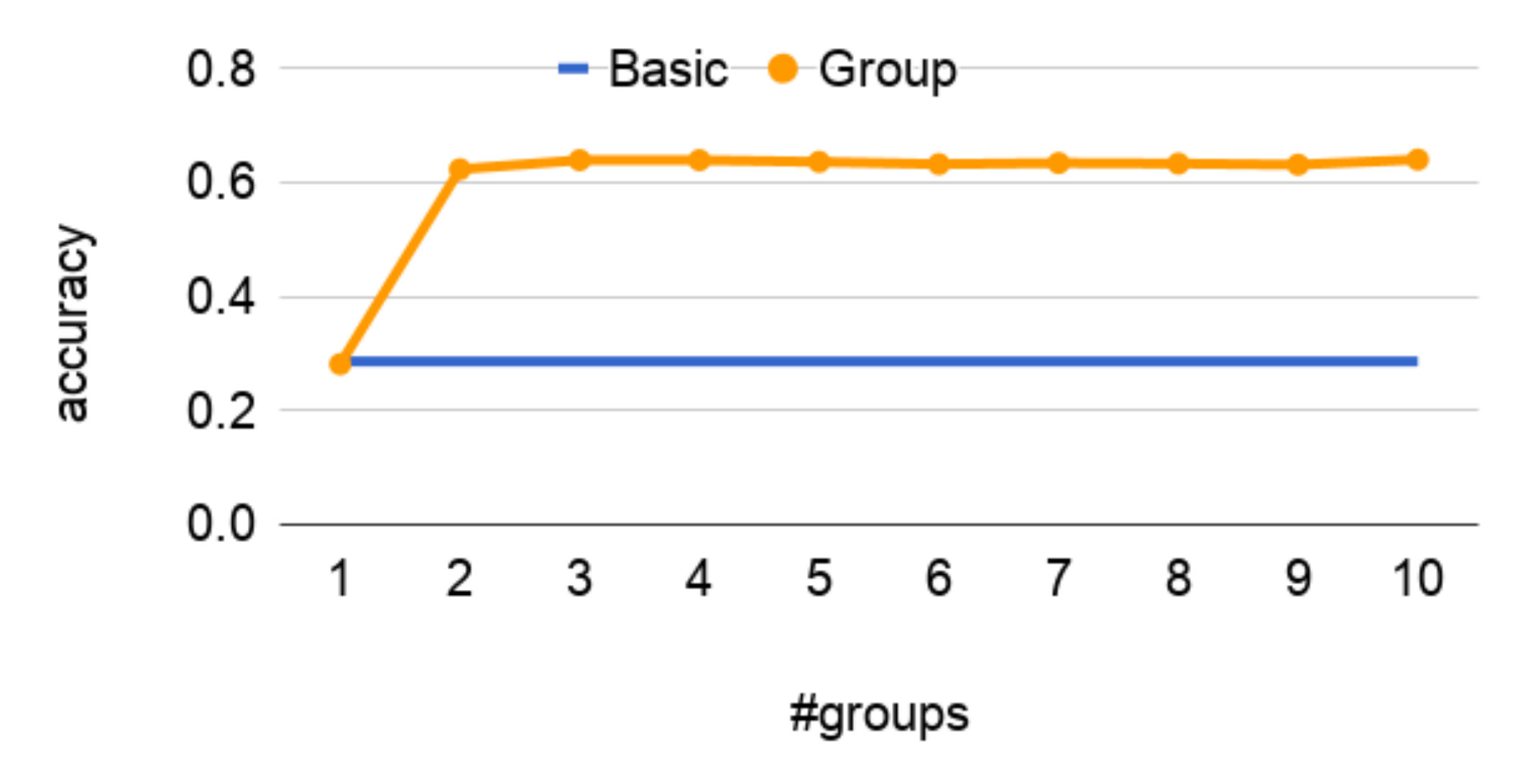} 
\vspace{-1cm}
\caption{Prediction accuracies in the comprehension experiment with synthetic data.} 
\label{fig:synth-acc} 
\end{figure}

\begin{figure}
\centering
\includegraphics[width=1.0\columnwidth]{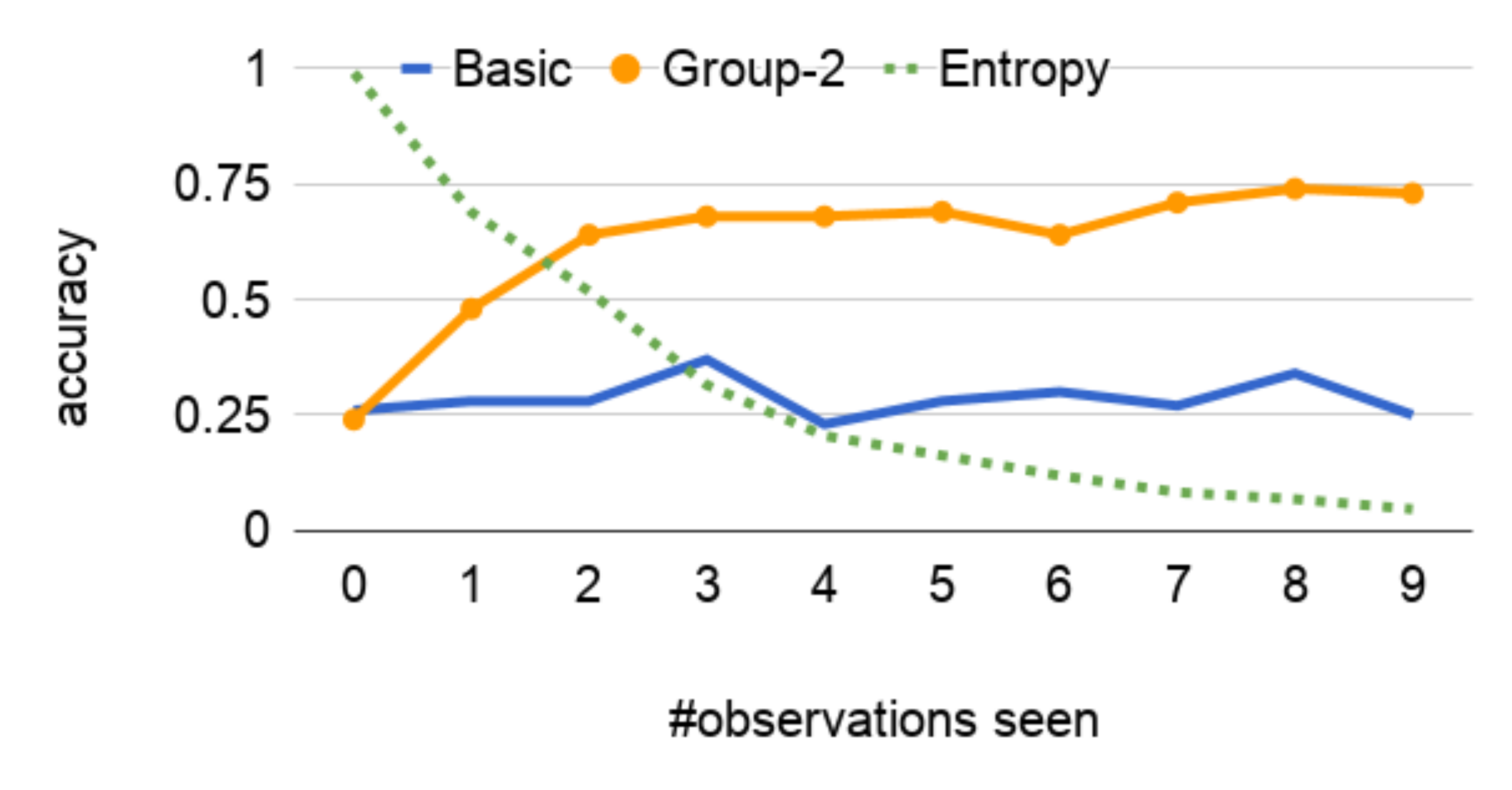} 
\vspace{-1cm}
\caption{Accuracy evolution with increasing number of observations from the user in the comprehension experiment with synthetic data.} 
\label{fig:synth-time} 
\end{figure}

We evaluated the same models in a 2-fold cross-validation. 
Figure~\ref{fig:synth-acc} shows
the prediction accuracy for \emph{Basic} and the \emph{Group} models for $K$ from 1 to 10. 
All models for $K>1$ clearly outperform the baseline model: the 2-group model gets
$62.3\%$ vs $28.6\%$ averaged over
all test examples, while adding more 
than two groups does not further improve the accuracy. 
We also show in Figure~\ref{fig:synth-time} the evolution of the accuracy as 
$D\up{u}$ grows: the \emph{Group} model with $K=2$ reaches a 64\% testing accuracy 
after seeing two observations from the same user. In the same figure,
the entropy of the posterior distribution over groups (see production experiment)
falls towards zero as $D\up{u}$ grows.
These results show that our model is capable of correctly assigning a user to the group 
they belong to, once the features are adequate for distinguishing between different 
user behaviors.

\subsection{Discussion}

Our model was shown to be successful in discovering groups 
of users with respect to their behavior, within datasets which present
discernible user variation. In particular, if all listeners are influenced in a similar way
 by e.g. the visual salience of an object, then the group model cannot 
learn different weights for the visual salience feature; if this happens 
for all available features, there are effectively no groups for our model to discover.

Once the groups have been discovered, 
our model can then very quickly distinguish between them at test time. 
This is reflected in the steep performance improvement even after the first
user observation in both the real data experiment in~\ref{sub:production}
and the synthetic data experiment in~\ref{sub:comprehension}.




\section{Conclusion}

We have presented a probabilistic model for NLG which predicts
the behavior of individual users of a dialog system by dynamically
assigning them to user groups, which were discovered
during training\footnote{Our code and data is available in \url{https://bit.ly/2jIu1Vm}}. 
We showed for two separate NLG-related tasks,
RE production and RE comprehension, how our model, 
after being trained with data that is not annotated with user groups,
can quickly adapt to unseen users as it gets more observations 
from them in the course of a dialog and makes increasingly 
accurate predictions about their behavior.

Although in this work we apply our model to tasks related to NLG,
nothing hinges on this choice; it can also be applied to any other 
dialog-related prediction task where user variation plays a role. 
In the future, we will also try to apply the basic principles of our 
user group approach to more sophisticated underlying models, such 
as neural networks.

\bibliography{mybib,sfb}
\bibliographystyle{acl_natbib}

\end{document}